\documentclass[runningheads]{llncs}
\usepackage{graphicx}
\usepackage{tikz}
\usepackage{listings}
\usepackage{xcolor}
\usepackage{makecell}
\usepackage{hyperref}
\usepackage{cite}
\usepackage{svg}
\usepackage{algorithm}
\usepackage[noend]{algpseudocode}
\usepackage{kmath}
\usepackage{textcomp}
\usepackage{pifont}
\usepackage{array}
\usepackage{enumitem}
\usepackage{soul}
\usepackage{marvosym}
\usepackage{ifsym}

\algrenewcommand\algorithmicrequire{\textbf{Input:}}
\algrenewcommand\algorithmicensure{\textbf{Output:}}

\colorlet{punct}{black}
\definecolor{delim}{RGB}{0, 0, 0}
\definecolor{Red}{RGB}{204, 0, 0}
\definecolor{OliveGreen}{rgb}{0.0, 0.5, 0.0}
\colorlet{numb}{black}

\hypersetup{
    colorlinks=true,
    linkcolor={OliveGreen},
    citecolor={OliveGreen},
    urlcolor={OliveGreen}
}

\lstdefinelanguage{json}{
    basicstyle=\small\ttfamily,
    showstringspaces=false,
    breaklines=true,
    frame=None,
    frameround=tttt,
    backgroundcolor=\color{white},
    literate=
     *{0}{{{\color{numb}0}}}{1}
      {1}{{{\color{numb}1}}}{1}
      {2}{{{\color{numb}2}}}{1}
      {3}{{{\color{numb}3}}}{1}
      {4}{{{\color{numb}4}}}{1}
      {5}{{{\color{numb}5}}}{1}
      {6}{{{\color{numb}6}}}{1}
      {7}{{{\color{numb}7}}}{1}
      {8}{{{\color{numb}8}}}{1}
      {9}{{{\color{numb}9}}}{1}
      {:}{{{\color{punct}{:}}}}{1}
      {,}{{{\color{punct}{,}}}}{1}
      {\{}{{{\color{delim}{\{}}}}{1}
      {\}}{{{\color{delim}{\}}}}}{1}
      {[}{{{\color{delim}{[}}}}{1}
      {]}{{{\color{delim}{]}}}}{1},
}

\begin{document}
\title{CCAE: A Corpus of Chinese-based Asian Englishes}
%
%
\author{Yang Liu\inst{} \and
Melissa Xiaohui Qin\inst{}\and
Long Wang\inst{} \and
Chao Huang\inst{}$^{(\textrm{\Letter})}$}
\authorrunning{Y. Liu et al.}
%
\institute{University of Science and Technology Beijing \\
\email{\{yangliu.real,dearmelissaqin\}}@gmail.com \\
\email{\{lwang,chaohuang\}@ustb.edu.cn}}
\maketitle            
\begin{abstract}
Language models have been foundations in various scenarios of NLP applications, but it has not been well applied in language variety studies, even for the most popular language like English. This paper represents one of the few initial efforts to utilize the NLP technology in the paradigm of World Englishes, specifically in creating a multi-variety corpus for studying Asian Englishes. We present an overview of the \textbf{CCAE} — \textbf{C}orpus of \textbf{C}hinese-based \textbf{A}sian \textbf{E}nglish, a suite of corpora comprising six Chinese-based Asian English varieties. It is based on 340 million tokens in 448 thousand web documents from six regions. The ontology of data would make the corpus a helpful resource with enormous research potential for Asian Englishes (especially for Chinese Englishes for which there has not been a publicly accessible corpus yet so far) and an ideal source for variety-specific language modeling and downstream tasks, thus setting the stage for NLP-based World Englishes studies. And preliminary experiments on this corpus reveal the practical value of CCAE. Finally, we make CCAE available at \url{https://huggingface.co/datasets/CCAE/CCAE-Corpus}.

\end{abstract}
\section{Introduction}

Natural language process (NLP) has achieved significant advances with the deep learning approaches in the domain of language modeling (LM), specifically the second-generation pre-trained language models (PLMs)\cite{Qiu2020PretrainedMF} such as BERT\cite{Devlin2019BERTPO}, T5\cite{raffel2020exploring}, and GPT-3\cite{brown2020language}, which are based on transformer backbone\cite{vaswani2017attention}. PLMs are fine-tuned to the target languages or the tasks at hand, so other researchers do not have to perform expensive pre-training. Due to their advanced generalization performance, PLMs have been utilized in a wide range of downstream applications, such as machine translation\cite{vaswani2017attention}, text classification\cite{minaee2021deep}, and question answering\cite{rajpurkar-etal-2018-know}. They have also been proved fruitful in capturing a wealth of linguistic phenomena and features on levels of morphology\cite{Edmiston2020ASA}, lexis\cite{espinosa-anke-etal-2021-evaluating, zhou-etal-2019-bert}, and syntax\cite{tran-bisazza-2019-zero}. Meanwhile, they can also be applied in relevant tasks such as variety detection\cite{zaharia-etal-2020-exploring} and lexical variation identification\cite{laicher-etal-2021-explaining}. 

\begin{figure}
    \centering
    \includegraphics[width=0.75\linewidth]{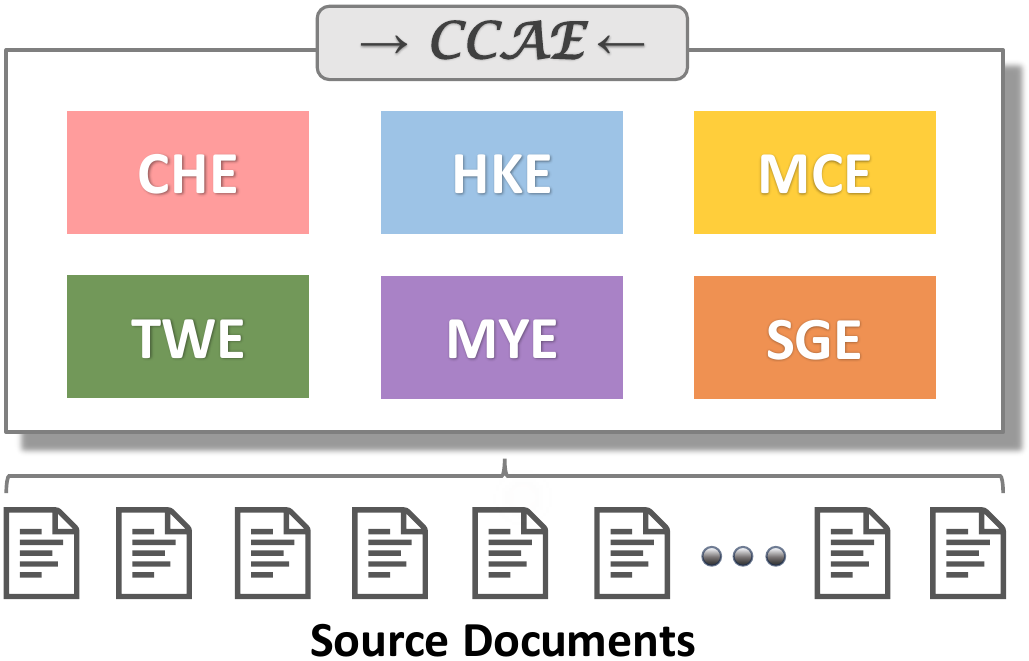}
    \caption{Components of CCAE, totally including six varieties.}
    \label{fig:corpus_components}
\end{figure}

World Englishes has become a robust field of inquiry as scholars pursue more nuanced understandings of linguistic localization and multilinguals’ negotiations of language differences\cite{nuske2018mean}. However, there have been few attempts to investigate various indigenized Englishes by means of PLMs. This study represents the initial effort to fill this gap by creating the first free-access supra corpus on which PLMs could be pre-trained for the Chinese and Chinese-based Asian English (CAE) varieties.  While previous corpora have been built for Inner and Outer Circle varieties such as the small structured ICE\cite{kirk2018international} and large-scale GloWbE\cite{davies2015expanding}, there has not been a publicly accessible corpus for the Expanding Circle English\cite{berns2005expanding}, Chinese English\cite{xu2020chinese}, Chinese influenced and Chinese based varieties such as Singapore English\cite{leimgruber2011singapore}. The corpus we are introducing is going to be an important data infrastructure for Asian Englishes study. \\
\indent In this paper, we present the CCAE (\textbf{C}orpus of \textbf{C}hinese-based \textbf{A}sian \textbf{E}nglishes), a suite of corpora totaling 340 million words in 448 thousand documents from six locations where Chinese-based English varieties were spoken. By Chinese-based Englishes, we mean Englishes developed in  Sinophone regions where varieties of Chinese are used as a main language of communication and thus serve as the dominating indigenous language or one of the indigenous languages  contacting English in the monolingual or multilingual contact setting. That is to say, Chinese is the dominating agent in the formation of the nativities Englishes or has influenced the formation of various localized Englishes. In order to form a definitive research scope, we include six regional varieties (Figure \ref{fig:corpus_components}) under the umbrella term of Chinese-based Englishes: Chinese mainland English (CHE), Hong Kong English (HKE), Macao English (MCE), Taiwan English (TWE), Malaysia English (MYE) and Singapore English (SGE).

The CCAE has the following major properties:
\begin{itemize}[noitemsep,topsep=0pt]
	    \item[$\bullet$] It is the first and largest open-access web-crawled corpus for the Chinese Englishes and Chinese-based Asian Englishes.
	    \item[$\bullet$] It is the first and largest NLP-oriented corpus for Asian Englishes and World Englishes.
	    \item[$\bullet$] It is a clean and deduplicated corpus in the document level. Taking into account the significance of data quality for dependent tasks, we introduce custom pipeline to conduct data cleaning.
        \item[$\bullet$] It maintains the traceability of each document to its origin. This level of traceability makes it possible for researchers to apply the withdrawal right of individual website owners or individual persons whose data are cited on websites and are protected by GDPR\cite{voigt2017eu}. It allows researchers to systematically exclude blacklisted websites.
        \item[$\bullet$] It serves as the initial data resource for potential usage on downstream tasks like language variety identification, lexical variation identification, and so on.
\end{itemize}

\section{Related Work}
As shown in Table \ref{table:CorporaComparison}, we compare CCAE with four other corpora. Here, we simply illustrate them\footnote{Note that whatever GloWbE, ICE or ACE, they are not NLP-oriented originally, and we only counted disk size, documents and tokens on the parts of CAE in them, separately.}, which is web-based or manually curated.

\begin{table}[ht]
\caption{CCAE versus other World English corpora \& WikiText. * stands for the unreported item in its bibliography, and - means ``not applicable". In addition, WikiText here refers to Wikitext-103.} 
\centering 
\begin{tabular}{p{26mm} >{\centering\arraybackslash}p{15mm} >{\centering\arraybackslash}p{24mm} >{\centering\arraybackslash}p{16mm} >{\centering\arraybackslash}p{18mm} >{\centering\arraybackslash}p{19mm}} 
\Xhline{2\arrayrulewidth} 
\textbf{Corpus}   & \textbf{GloWbE} & \textbf{ICE}       & \textbf{ACE}  & \textbf{WikiText} & \textbf{CCAE (ours)} \\ [0.5ex] 
\hline 
Varieties (CAE)   & 3               & 2                  & 5             & 0                 & 6     \\
Disk Size         & 686MiB          & 400MiB             & 2.1MiB        & 500MiB            & 2.2GiB \\
Documents         & 134k            & -                  & -             & 23.8k             & 448k  \\
Tokens            & 142M            & 1.8M               & 420k          & 100M              & 340M  \\
Parsing Quality   & Low             & High               & High          & High              & High  \\
Cleaning Quality  & Low             & High               & High          & High              & High  \\
Corpus Type       & Web             & Spoken \& Written  & Spoken        & Web               & Web   \\
Rich Metadata & \ding{51}       & \ding{51}          & \ding{55}     & \ding{55}         & \ding{51}  \\
Deduplicated      & \ding{55}       & \ding{51}          & \ding{51}     & \ding{51}         & \ding{51}   \\
Open Licence      & \ding{55}       & \ding{55}          & \ding{55}     & \ding{51}         & \ding{51} \\
\Xhline{2\arrayrulewidth} 
\end{tabular}
\label{table:CorporaComparison} 
\end{table}

\noindent\textbf{GloWbE.}\ \ The corpus of Global Web-based English is a large-scale collection of 1.8 million web pages from 20 English-speaking countries, containing over 1.9 billion tokens. It provides linguistic annotations like PoS to support the investigation of how English is used globally. 

\noindent\textbf{ICE.}\ \ The International Corpus of English is a collection of spoken and written English from 20 regions where English is used as the first or second language. It includes over 1,000 texts and 1,200 hours of audio recordings, making it a valuable resource for studying varieties of English language use across regions and cultures around the world.

\noindent\textbf{ACE.}\ \ The Asian Corpus of English \cite{kirkpatrick2013asian}, an Asian English-oriented corpus capturing spoken ELF (English as a lingua franca) interactions in various regions of Asia.

\noindent\textbf{WikiText-103.}\ \ WikiText-103 \cite{merity2016pointer} consists of over 100 million tokens extracted from English Wikipedia, it is commonly used as a benchmark dataset for training and evaluating language models. This corpus can be deemed as one of the representations of Inner-circle English.

\section{CCAE at a Glance}

To comprehend accurately, it is essential to understand the origin of the texts that form it. Therefore, we describe CCAE's text and metadata respectively in terms of (1) corpus-level statistics, (2) the frequency of various internet domains as text sources, and (3) the utterance date when the websites were initially indexed.

\begin{table}[ht]
\caption{Corpus-level statistics for CCAE.} 
\centering 
\begin{tabular}{>{\raggedright\arraybackslash}p{10mm} >{\raggedleft\arraybackslash}p{15mm} >{\raggedleft\arraybackslash}p{15mm} >{\raggedleft\arraybackslash}p{15mm} >{\raggedleft\arraybackslash}p{12mm} >{\raggedleft\arraybackslash}p{12mm} >{\raggedleft\arraybackslash}p{30mm} >{\raggedleft\arraybackslash}p{22mm}} 
\Xhline{2\arrayrulewidth} 
\textbf{Variety}   & \textbf{Disk Size}    & \textbf{Weight}    & \textbf{Websites} & \textbf{Docs}   & \textbf{Tokens}  & \textbf{Mean Document Size} \\ [0.5ex] 
\hline 
CHE       & 766MiB   & 33.39\%    & 145k     & 147.3k & 114M   & 5.32KiB \\ 
HKE       & 410MiB   & 17.87\%    & 90k      & 90.5k  & 62M    & 4.63KiB \\
MCE       & 33MiB    & 1.44\%     & 9k       & 9.3k   & 5M     & 3.63KiB \\
TWE       & 307MiB   & 13.38\%    & 46k      & 46k    & 42M    & 6.83KiB \\
MYE       & 258MiB   & 11.25\%    & 51k      & 51.5k  & 40M    & 5.12KiB \\
SGE       & 520MiB   & 22.67\%    & 103k     & 103.3k & 77M    & 5.15KiB \\
\hline
TOTAL     & 2.2GiB   &            & 438k      & 448k  & 340M & 5.24KiB \\
\Xhline{2\arrayrulewidth} 
\end{tabular}
\label{table:CorpusStatisics} 
\end{table}

\subsection{Corpus-level Statistics}

We collected a total of 101GB WARC(Web ARChive)\footnote{See the following Wikipedia page for more information on this standard file format:\newline\url{https://en.wikipedia.org/wiki/Web_ARChive}} files for the CCAE. After document-level deduplication, the corpus is composed of 448k documents and 340M word tokens(measured by SpaCy\footnote{SpaCy Tokenizer: \url{https://spacy.io/api/tokenizer}} tokenization). Basic statistics of the disk size for the cleaned corpus, collected websites, documents, and tokens are displayed in Table \ref{table:CorpusStatisics}.

\begin{figure}[htp!]
    \centerline{\includegraphics[width=1.18\linewidth]{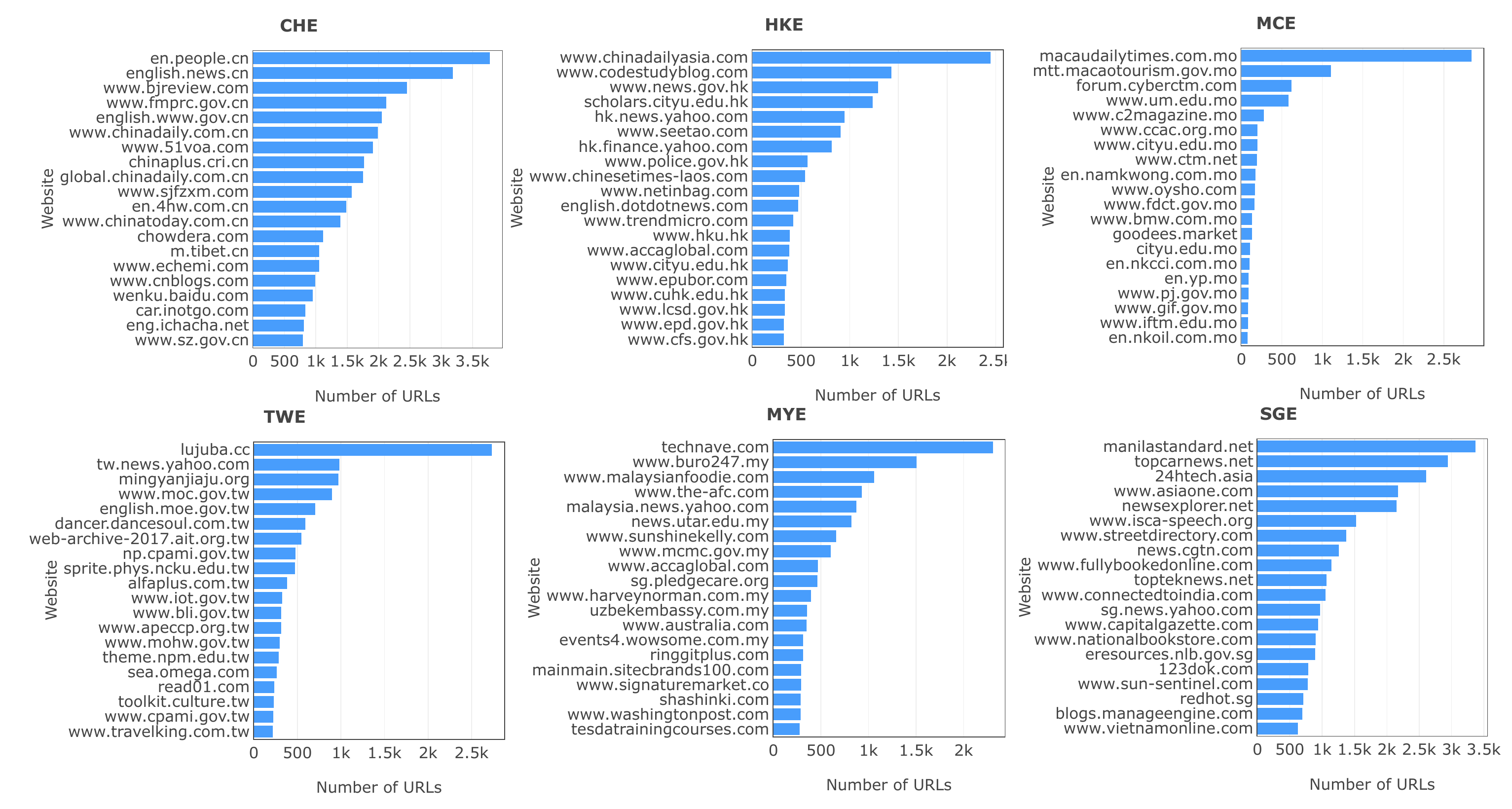}}
    \caption{Top 20 highest frequently occurring websites for each variety.}
    \label{fig:domain_frequency}
\end{figure}

\subsection{Domains Distribution}
We have conducted analysis on the highest frequent top-level domains (TLD) for each variety. Predictably, most of the represented URLs are from some popular top-level domains like .com, .net, and .org. Apart from this common case, the URLs mainly consist of variety-corresponding TLD, for instance, “Chinese Mainland” has nearly 57\% portion for “.cn”, and “Hong Kong” has 34\% portion for “.hk”.

In addition, we present the top 20 highest frequently occurring websites for each variety in Figure \ref{fig:domain_frequency}, to display the distribution of text across different websites for each variety.

\subsection{Utterance Date}
Language undergoes change quickly, and the accuracy of statements depends on when they were made. We attempted to determine the date of each document by examining the publish date from two sources: Google search and Internet Archive\footnote{Internet Archive: \url{https://archive.org/web}}. We used the earlier date as the publish date for each web page. We note that the use of the Internet Archive is not perfect, as it sometimes indexes pages months after their creation and only indexes around 65\% of the URLs in CCAE. For web pages with unknown dates, we marked them as ``NULL" in later storage.

\begin{figure}[htp!]
\hspace{-0.5cm}
\includegraphics[width=1.2\linewidth]{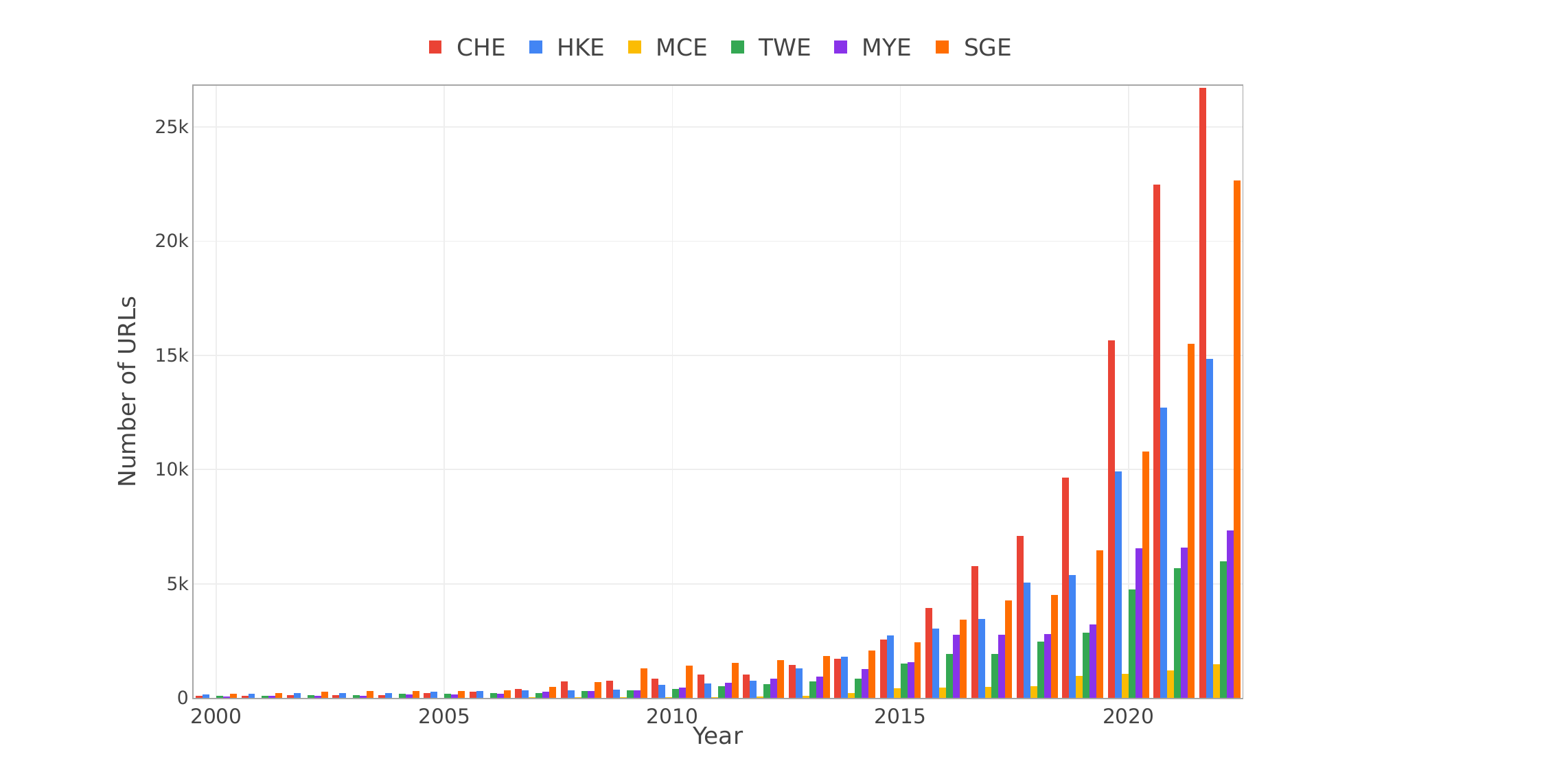}
\caption{The date when URLs were first indexed by the Google or Internet Archive in six Asian English Varieties}
\label{fig:data_index_date}
\end{figure}

As shown in Figure \ref{fig:data_index_date}, regardless of variety, we found that the dates of approximately 96\% URLs were distributed from 2011 to 2022. In addition, there is also a significant amount of data that was written 10 years before the data collection period (from 2022/01 to 2022/06), indicating a long-tailed distribution.

\section{Generation of CCAE}
\subsection{Data Collection}
To create a corpus available to permit research on a wide range of phenomena in six Chinese-based Englishes by performing downstream tasks in NLP (or conventional linguistic approaches), there are three principal considerations: (a) major distribution coverage, (b) variety accuracy (to ensure that the web pages were correctly associated with each of the six locations) and (c) domain diversity.

\textbf{Major distribution coverage} is crucial for creating corpora that can yield rich language features and be beneficial to models' generalization. To achieve this, we used hundreds of the highest-frequency trigrams as initial keyword seeds for query in COCA \cite{davies2010corpus}. These trigrams include common three-word phrases such as "one of the" and "as well as". Algorithm \ref{alg:algo1} shows the procedure of generating a query set. We believe that using the most frequently occurring trigrams as queries is a better approach while retrieving on Google, rather than randomly selecting keywords, as it better represents the natural data distributions across different domains.

\begin{algorithm}
\caption{Generate trigrams as query set}\label{alg:algo1}
\begin{algorithmic}[1]
\Require Corpora\ \textbf{$\mathcal{C}$} = $\left\{d_i\right\}$, i = 1, ..., Sizeof($\mathcal{C}$)
\Ensure  Trigram\ Query\ Set\ $\textbf{$\mathcal{S}$}$
\State $\mathcal{L}$ $\leftarrow$ List()
\State $\mathcal{D}$ $\leftarrow$ Dictionary()
\For{$d_i\ \in\ \mathcal{C}$} \Comment{iterate each document with fixed window size $\equiv$ 3}
\State \textit{l}, \textit{r} $\leftarrow$ 0, 3;
\State $\mathcal{L}$ $\leftarrow$ $t$ $\in$ Tokenize($d_i$) \Comment{tokenize each document by whitespace}
\While{\textit{r} $<$ Sizeof($\mathcal{L}$)}
    \State gram $\leftarrow$ $``\ "$.Join($\mathcal{L}$[\textit{l}:\textit{r}]) \Comment{splice tokens in the window [\textit{l}, \textit{r}) with whitespace}
    \If{gram\ not\ in\ $\mathcal{D}$}
        \State $\mathcal{D}$[gram] $\leftarrow$ 0
    \EndIf
    \State $\mathcal{D}$[gram] $\leftarrow$ $\mathcal{D}$[gram] + 1
    \State \textit{l} $\leftarrow$ \textit{l} + 1
    \State \textit{r} $\leftarrow$ \textit{r} + 1
\EndWhile
\EndFor
\State Sort($\mathcal{D}$) \Comment{sort $\mathcal{D}$ by value}
\State \textbf{$\mathcal{S}$} $\leftarrow$ GetTopK($\mathcal{D}$) \Comment{get the top-k most frequent trigrams as query set $S$}
\end{algorithmic}
\end{algorithm}

\textbf{Variety accuracy} is also important to ensure that web pages are associated correctly with each of the six regions in the corpus. To achieve that, we run the trigram list we have generated against Google advanced search, in specific “search region”, corresponding to each one of the varieties in China Mainland, Hong Kong, Macao, Taiwan, Malaysia, and Singapore. This method is shown to be credible \cite{davies2015expanding} as Google search has adopted the following policies on web page crawling: (1) recognizing the top-level domain region (e.g., \textit{.cn} for China Mainland, \textit{.hk} for Hong Kong); (2) Identifying the web server's IP address; (3) Determining who links to the website; and (4) Analyzing the website's visitors, which has correctly identified website by its region. Through this method, we collected all the URLs for each item (i.e. a trigram as a query), in the result page and finally generate a deduplicated URL set for each variety.

\textbf{Domain diversity} is a very important factor for the representativeness of the corpus. Collected documents should cover as many domains as possible such as technology, sport, finance, and arts. By employing human checking manually, in detail, it involves employing annotators to manually verify and validate the collected data. The annotators ensure that the collected data represents a diverse range of domains. This helps to confirm that the resulting dataset is balanced and representative of the language being studied. 

We developed a collector to leverage Selenium\footnote{Selenium: \url{https://www.selenium.dev}} and ChromeDriver\footnote{Webdriver for chrome: \url{https://chromedriver.chromium.org}} to simulate human behavior, for collecting URLs. To address the obstruction of reCAPTCHA — the anti-crawled system which Google search adopts, we use 2captcha\footnote{2captcha - a captcha solving service: \url{https://2captcha.com}}, a third-party online service for bypassing its verification code. To balance the requested servers' workload, we used different request proxies from around the world and keep the query frequency low to be friendly to response servers.

After the creation of the URL set for each variety, the final step of the collection is to start the downloader script to download the web page corresponding to each URL. Subsequently, we generate WARC files that contain every request and response we sent, this allows us to experiment with later processing, without hitting the server again. The crawling exclusively scraped websites whose robot files allowed it, resulting in a total of 438,625 websites across six varieties. The final raw crawling size is about 101GB of WARC files.

\subsection{Data Pre-processing}
Data quality is key when building a corpus. In this section, we discuss the corpus’ data pre-processing from three aspects including parsing, cleaning, and deduplication. We introduce our pipeline to accomplish the tasks of pre-processing.

In general, a web page contains different parts (e.g., header, body, tail), while we only need text body. To extract text on accuracy concern, we introduce a web text extraction tool JusText \cite{pomikalek2011removing} which is based on a heuristic approach, to extract text from HTML pages.

After running JusText, we filter out any lines that don't end with a terminal punctuation mark. This is to guarantee that the sentences that stay are both valid and meaningful. Documents with less than five sentences are eliminated as they may not provide enough context to comprehend the text. Additionally, any unnecessary symbols or punctuations are taken out of the text, unsuitable words or words that are deemed offensive are also removed.

We noted that some of the URLs are wrongly associated with their respective regions, for example, a URL with “hk” turned out to be identified with the region of Macao (appears in the result of Macao collections). We guess it is possible as there is a quite low error rate of archive events in Google index policy, even if it follows almost credible strategies of web page categorization in the above discussion. We moved misidentified URLs to their correct class. 

\subsection{Output Storage Format}
The output data format of CCAE we defined is JSON, there is a unique JSON document with the following data fields:

\begin{itemize}
  \item[$\bullet$] TextID: unique document identifier consisting of an eight-digit-width integer over the whole corpus.
  \item[$\bullet$] Time: this field of data can be used to track changes in the variety of language use over time and to identify trends in language variation.
  \item[$\bullet$] Words: word count of this document.
  \item[$\bullet$] Variety: English variety this document belongs to.
  \item[$\bullet$] URL: URL address from which the web page content has been extracted.
  \item[$\bullet$] Title: textual title of this document. 
  \item[$\bullet$] Content: full pre-processed text of this document.
\end{itemize}

A sample JSON document in the corpus is shown below.

\begin{center}
\begin{minipage}{.9\textwidth}
\begin{lstlisting}[language=json, firstnumber=1]
{
    "TextID": 00019734,
    "Time": "2019-01-28",
    "Words": 741,
    "Variety": "cn",
    "Genre": "G",
    "Domain": "www.scyxxc.com",
    "URL": "http://www.scyxxc.com/en/m/news/370.html",
    "Title": "<textual title>",
    "Content": "<textual content>"
}
\end{lstlisting}
\end{minipage}
\end{center}

As for the text with linguistic tags, we provide another version of data to support it, each word in this replica is aligned with its multiple tags like lexeme and PoS.

\section{Applications of CCAE}
In this section, we demonstrate how CCAE can be used for tasks like variety-oriented language modeling and automatic variety identification, and discuss its usage for further research. We note that the utility of CCAE stretches beyond the two use cases to a wider range of language variety-relevant text mining tasks.
\subsection{Multi-variety Language Modeling}
\subsubsection*{Task.}As one of the few trials to combine NLP with World English, we conduct a preliminary experiment on the task of language modeling. We investigate different experimental settings on multi-variety Asian English through perplexity computations by GPT-2 \cite{radford2019language}. Through the experiments, we hope to shed light on the unique linguistic characteristics of multi-variety Asian English and the challenges it presents for language modeling. 

\begin{table}[ht]
\caption{Zero-shot (ZS) \& Fine-tuning (FT) performance for six varieties. We evaluate test perplexity (lower is better) on the validation set, for each variety, we use its abbr. to refer to itself.} 
\centering 
\begin{tabular}{p{30mm} >{\centering\arraybackslash}p{12mm} >{\centering\arraybackslash}p{12mm} >{\centering\arraybackslash}p{12mm} >{\centering\arraybackslash}p{12mm} >{\centering\arraybackslash}p{12mm} >{\centering\arraybackslash}p{12mm} >{\centering\arraybackslash}p{9mm} >{\centering\arraybackslash}p{15mm}} 
\Xhline{2\arrayrulewidth} 
\textbf{CCAE}     & \textbf{CHE}     & \textbf{HKE}     & \textbf{MCE}      & \textbf{TWE}     & \textbf{MYE}     & \textbf{SGE}     & \textbf{Avg} \\ [0.1ex]
\hline
GPT-2-ZS                    & 21.2     & 21.0    & 20.6     & 32.2    & 28.4    & 21.4    & 24.1 \\ [1pt]
GPT-2-FT (mixture)          & 16.2     & 16.1    & 15.2     & 24.9    & 17.9    & 15.9    & 17.7 \\ [1pt]
GPT-2-FT (specific)          & \textbf{15.6}     & \textbf{15.0}    & \textbf{12.1}     & \textbf{24.3}    & \textbf{15.2}    & \textbf{14.8}    & \textbf{16.1} \\ [1pt]
\Xhline{2\arrayrulewidth} 
\end{tabular}
\label{table:SFTGPT} 
\end{table}

\subsubsection*{Setup.}The baseline model used in our experiment is the 345M GPT-2 model, implemented by Hugging Face Transformers \cite{wolf2020transformers} and PyTorch \cite{paszke2019pytorch}. We utilize Adam optimization \cite{kingma2014adam} with $\beta_1$ = 0.9, $\beta_2$ = 0.999, set the dropout \cite{srivastava2014dropout} value to 0.1, and use a learning rate of 5e-5 with batch size of 65,536 tokens. Training is conducted for a maximum of 100k iterations, with early stopping performed over the validation set. The experiments are run on 8 $\times$ NVIDIA A100-80GB GPUs.

Respectively, we conduct three settings of runs: (1) Zero-shot prompting: we simply drive the model on the validation set which we split, without any training; (2) Fine-tuning with mixed training sets across six varieties: we fine-tune the models with data of each specific English variety and test it with corresponding validation set and (3) Fine-tuning with training set of each specific variety: we merge all the parts of the training set of each variety, and training one model, and then test the model on the original split of validation set for each variety. 

\subsubsection*{Results and discussion.}As shown in Table \ref{table:SFTGPT}, We first use the original checkpoint of GPT-2 to compute the token level perplexity on each English variety directly. And then we carry out the same evaluation with the setting of supervised fine-tuning (SFT) \cite{ouyang2022training}. Intuitively, the results show that the data from CCAE can significantly improve the language modeling performance on the metric of perplexity. We argue that specific SFT has considerable potential to increase the capability of ``understanding" different language varieties, compared with fine-tuning with mixture of data. It indicates that the necessity of creations for variety-aware language models are nonnegligible. However, delicate experiments still need to be designed to consider the impact of variant varieties on language models with SFT in order to obtain a final credible conclusion.

\subsection{Automatic Variety Identification}
\subsubsection*{Task.}Automatic variety identification (AVI) is a more intricate and nuanced task compared to language identification, as it demands the capability to differentiate between numerous variations of a single language, and the linguistic variations among related varieties are less apparent than those among distinct languages\cite{yang-xiang-2019-naive}. Consequently, it has become an appealing subject for many researchers in recent years \cite{popa-stefanescu-2020-applying, ceolin-2021-comparing}.

\begin{table}[ht]
    \caption{Precision, Recall and F1 for variety identification experiment on validation set.} 
    \centering 
    \begin{tabular}{>{\centering\arraybackslash}p{17mm} >{\centering\arraybackslash}p{13mm} >{\centering\arraybackslash}p{11mm} >{\centering\arraybackslash}p{11mm} >{\centering\arraybackslash}p{11mm}} 
    \Xhline{2\arrayrulewidth} 
    \textbf{Variety}       & \textbf{Precision}   & \textbf{Recall}   & \textbf{F1}     & \textbf{\#}     \\ [0.5ex] 
    \hline 
    CHE           & 80.46       & 80.02    & 80.24  & 1,472  \\ [0.5pt] 
    HKE           & 62.71       & 65.96    & 64.29  & 905   \\ [0.5pt]
    MCE           & 77.92       & 63.82    & 70.17  & 94    \\ [0.5pt]
    TWE           & 70.53       & 66.08    & 68.23  & 460   \\ [0.5pt]
    MYE           & 70.86       & 69.76    & 70.31  & 516   \\ [0.5pt]
    SGE           & 74.33       & 75.41    & 74.86  & 1,033  \\ [0.5pt]
    \hline
    macro avg     & 72.80       & 70.18    & 71.35  & 4,480  \\
    weighted avg  & 73.28       & 73.16    & 73.19  & 4,480  \\
    \Xhline{2\arrayrulewidth} 
    \end{tabular}
    \label{table:VarietyIdentification} 
\end{table}

\subsubsection*{Setup.}CCAE naturally supports the task of AVI, thanks to its rich metadata. We present our preliminary trial on the AVI task(essentially, it is a long document classification task for our data) for six Asian English varieties in CCAE. In brief, we employ Longformer \cite{beltagy2020longformer} and fine-tune it as an examplar baseline on our dataset, which is random-sampled in the proportion of the original distribution from CCAE. The label of each datapoint is generated by its original variety type, after carrying out a manual inspection on a randomly selected sample, we determined the precision of the label confidence, which resulted in \textgreater0.95. This validates the caliber of our supervised information and, consequently, our resources.

\subsubsection*{Results and discussion.}Not surprisingly, the results of this experiment (cf. Table \ref{table:VarietyIdentification}) clearly highlight that the few-example categories seem to be more difficult to capture the unique characteristics of its class. To advance significantly, it suggests that compiling larger datasets and collecting more examples for minority classes (especially for MCE), or employing more advanced models to solve this problem are considerable optimizatized directions.

\section{Conclusion and Future Work}
We develop CCAE, a novel multi-variety web corpora for enhancing Asian Englishes study. CCAE contains six English varieties in Asian Chinese-speaking areas. The corpus provides affluent metadata and annotations for usages. We host two versions of the data for download, to spur further research on building language variety-adaptive language models on our corpus.

In future work, we suggest an in-depth investigation of variety-specific downstream tasks like multi-variety language modeling and automatic variety identification. Conduct experiments using CCAE to analyze the effectiveness of domain adaptation between various varieties. Through our work, we hope to encourage the community to further study World Englishes, to boost non-biased and culture-diversified language modeling development.

\section{Acknowledgement}
 Many thanks to Mark Davis, for his useful suggestions on data collection. We also thank the Internet Archive for providing service on the website time archive. This work was supported in part by the National Natural Science Foundation of China under Grant 62002016 and in part by the Fundamental Research Funds for the Central Universities under Grant 06500103.

\section{Data Availability}
CCAE has been released under the CC-BY-NC-ND 4.0\footnote{\url{https://creativecommons.org/licenses/by-nc-nd/4.0}} license on Hugging Face's website: \url{https://huggingface.co/datasets/CCAE/CCAE-Corpus}, enabling reusers to copy and distribute the material in its unadapted form, only for noncommercial purposes, while giving attribution to the creator.

%
%
%
%

\end{document}